# Alzheimer Disease Classification through ASR-based Transcriptions: Exploring the Impact of Punctuation and Pauses

*Lucía Gómez-Zaragozá[1], Simone Wills[2], Cristian Tejedor-Garcia[2], Javier Marín-Morales[1], Mariano Alcañiz[1], Helmer Strik[2]*

[1] Instituto Universitario de Investigación en Tecnología Centrada en el Ser Humano, Universitat Politècnica de València, Spain
[2] Radboud University, The Netherlands

`lugoza@i3b.upv.es, simone.wills@ru.nl, cristian.tejedorgarcia@ru.nl, jamarmo@i3b.upv.es, malcaniz@i3b.upv.es, helmer.strik@ru.nl`

## Abstract

Alzheimer's Disease (AD) is the world's leading neurodegenerative disease, which often results in communication difficulties. Analysing speech can serve as a diagnostic tool for identifying the condition. The recent ADReSS challenge provided a dataset for AD classification and highlighted the utility of manual transcriptions. In this study, we used the new state-of-the-art Automatic Speech Recognition (ASR) model Whisper to obtain the transcriptions, which also include automatic punctuation. The classification models achieved test accuracy scores of 0.854 and 0.833 combining the pretrained FastText word embeddings and recurrent neural networks on manual and ASR transcripts respectively. Additionally, we explored the influence of including pause information and punctuation in the transcriptions. We found that punctuation only yielded minor improvements in some cases, whereas pause encoding aided AD classification for both manual and ASR transcriptions across all approaches investigated.

## 1. Introduction

Alzheimer's Disease (AD) is a progressive neurodegenerative disorder characterised by a decline in cognitive functioning, with notable deterioration in memory, thought, and language [1]. It is the most common form of dementia, which affects an estimated 55 million people globally [2]. Despite the importance of early identification and intervention for the management of the disease, around 75% of cases go undiagnosed [2].

In the past few years, speech-based methods of AD classification, which leverage the speech abnormalities associated with AD, have proven promising at delivering an accurate, sensitive, and non-invasive means of automatically screening large populations for indications of Alzheimer's dementia [3]. Both the audio signal (acoustic features) and speech content (linguistic features) of spontaneous speech have been used in training AD classifiers using machine learning algorithms and deep learning approaches. In the recent Alzheimer's Dementia Recognition through Spontaneous Speech (ADReSS) challenge [4], systems based on linguistic features extracted from manual transcriptions were shown to outperform those trained solely on acoustic features. Moreover, the inclusion of para-linguistic features, such as pauses and disfluency markers, was proven to boost model performance [5, 6, 7, 8, 9].

In the machine learning approaches, of which Support Vector Machines (SVM) and k-Nearest Neighbors (kNN) were two of the most popular algorithms, the most prominent linguistic features were those carrying lexical, semantic and syntactic information. While there was a notable use of transformer-based word embeddings in deep learning approaches [5, 6, 7, 9, 10, 11, 12, 13, 14, 15]. As classification of AD using linguistic features relies heavily on the fidelity of the speech transcriptions, manual transcriptions offer the best possibility for training top performing classifiers. This is due to their high degree of accuracy and the ability to encode paralinguistic features within the text. Nevertheless, the use of manual transcriptions is undesirable due to its prohibitive time and resource costs, in addition to poor scalability.

The recent advances in ASR technologies, such as wav2vec2.0 [16] and Whisper AI [17], have made ASR-based transcriptions an increasingly viable option for use in AD classification, even considering the frequent use of spontaneous speech. As such, in this paper we investigate whether AD classification using speech content extracted from ASR-based transcriptions can perform comparatively to classification using manual transcriptions. We use the ADReSS dataset as it is one of the most recent challenges to provide a balanced dataset which includes manual transcriptions. Few attempts of using ASR-based transcriptions on the challenge dataset have been reported in literature [18, 19].We use the new state-of-the-art ASR model, Whisper, to obtain the ASR-based transcriptions. Unlike the previous widely used model, wav2vec2.0, Whisper produces transcriptions with punctuation marks included. We make a novel contribution by exploring the influence of the automatic punctuation marks on classification results, in addition to pause encoding. Particularly, we implemented pause encoding using the timestamp outputs of the WhisperX library, while the only previous study [19] that incorporated pause encoding to ASR transcriptions used wav2vec2-base-960h model for ASR and required an additional model to encode the pauses. We consider both machine learning and deep learning approaches, using linguistic features extracted by the Linguistic Inquiry and Word Count (LIWC) text analysis program [20] for machine learning and word embeddings as part of a deep learning approach. In the present paper we report on a study we conducted to pursue this research goal. The research questions we addressed are:

- *RQ1.* To what extent are ASR-based transcriptions useful for AD classification compared to manual transcriptions?
- *RQ2.* To what extent does the use of deep learning outperform traditional machine learning approaches for AD classification?
- *RQ3.* To what extent does the inclusion of punctuation and pauses in transcripts aid in AD classification?

The paper is structured as follows; in Section 2 we present the data and transcription generation process. Section 3 follows with a detailed description of the the classifier models used and the encoding of pause and punctuation information. Experiment results are reported in Section 4, before findings are discussed in relation to our research questions, in Section 5. The paper concludes with a brief overview of the work we have presented.

## 2. Data

This study was conducted using the dataset provided in the ADReSS challenge [4]. It consists of recordings of healthy controls (HC) and participants with AD describing the Cookie Theft picture from the Boston Diagnostic Aphasia Examination [21]. The training set includes 108 speakers (2 hours of audio utterances) and the test set 48 speakers (1 hour of audio utterances), both balanced in terms of gender, age and HC/AD participants. The ADReSS data includes the participant's speech recordings and the corresponding manual transcriptions. Further details of the dataset are described in [4].

The following subsections describe the processing applied to ADReSS data to obtain the two main data inputs used in this study: manual and ASR-based transcriptions, and their corresponding word-level timestamps.

### 2.1. Manual Transcriptions

The manual transcriptions given in the ADReSS challenge were annotated using the Codes for the Human Analysis of Transcripts (CHAT) format [22]. This coding system includes standardised ways to transcribe conversations and include meta-information related to the participant's behaviour, such as spelling mistakes, false word starts, noises (e.g. laughs, grunts, coughs) and filler sounds (e.g. mm, uh).

A first processing step was applied to keep the text corresponding to what the participant pronounced in his speech. Untranscribed words, out of vocabulary terms and anonymisation tags were mapped to "unk". Spelling mistakes, shortened words and phonetically spelled words were replaced by the regular form. Symbols representing additional information and comments were removed. Noises and incomplete/unintelligible terms were also eliminated but filler sounds were retained.

To obtain word-level timestamps, a Kaldi-based forced aligner using NNet2 online acoustic models was used to align processed transcriptions with the audio recordings.

### 2.2. ASR-based Transcriptions

The recordings in the ADReSS dataset contain elderly speech with background noise in some cases. Therefore, we decided to use different state-of-the-art ASR models to determine which one achieved the best transcriptions for this particular dataset: Kaldi ASpIRE[1], Wav2Vec2 [16] and Whisper [17].

The Kaldi ASpIRE Chain Model was released in 2016 by Kaldi [23]. It uses deep neural networks (DNNs) trained with the chain training criterion on Fisher English dataset, augmented with impulse responses and noises to create multi-condition training. The DNNs use a combination of time-delay neural networks (TDNNs) and bidirectional long short-term memory (BLSTM) networks to model the spectral and temporal variations in the speech signal.

The Wav2Vec2.0 model [16], released in 2020, is a transformer-based model which is first pre-trained using un-labelled audio only (called self-supervised learning) and then fine-tuned on transcribed speech. In this work, we used the cross-lingual approach, which learns speech representations by pre-training the model using audios in multiple languages. Specifically, we used the fine-tuned version on English of the Facebook's models wav2vec2-xls-r-1b[2] and wav2vec2-large-xlsr-53[3], both available on HuggingFace.

The Whisper ASR system, recently released by OpenAI, is a encoder-decoder transformer model pre-trained using 680,000 hours of multilingual data and a multi-task approach for both speech recognition and speech translation. Without the need of specific fine-tuning, Whisper generates very accurate transcriptions which also include punctuation. Moreover, the model provides utterance-level timestamps but word-level timestamps can be obtained using WhisperX Python library [24], which implements forced alignment using phoneme-based ASR models. We used both the base[4] and large[5] versions of Whisper model available on HuggingFace.

## 3. Methods

This section is divided in three parts. First, the modelling approaches applied to the AD classification are described in Section 3.1. Then, to study the influence of two factors on the classification results, the addition of pause information in the modelling approaches is detailed in Section 3.2, and the inclusion of punctuation marks in the transcriptions in Section 3.3.

### 3.1. Modelling

Two different modelling approaches were tested. The first based on feature engineering and classical machine learning algorithms, and the second based on word embeddings and deep learning models, explained in Section 3.1.1 and 3.1.2, respectively. The classification models were trained and tested using the corresponding data partitions provided in the ADReSS challenge. The results were evaluated using the following metrics: accuracy, precision, recall and F1 score. Their calculation is detailed in the challenge baseline article [4]. Implementation details can be found in the accompanying GitHub repository[6].

#### 3.1.1. Feature Extraction and Machine Learning

Although deep learning approaches are currently the state-of-the-art for text classification, we decided to also use a classical pipeline based on feature engineering (FE) followed by machine learning (ML) models because it is explainable, less computationally expensive and achieves good results when few data are available, as in this case. First, the LIWC feature set [20] was extracted for each participant's transcriptions using the LIWC2015 Dictionary version for English. This dictionary includes a comprehensive list of words and phrases organised into over 90 categories related to grammar, cognitive processes, emotional content and informal language, among others (see description in [25]) by means of the percentage of words belonging to each category with respect to the total number of words in the transcriptions analysed. LIWC features have been previously used for AD classification [26, 27, 28]. The features were standardised and the importance of each feature for clas-sification was determined by means of recursive feature elimination in order to obtain a rank-ordered list. For one less feature each time, the following two machine learning models were trained: Support vector machines (SVM) and k-nearest neighbours (KNN). Leave-One-Subject-Out cross-validation was applied to select the best hyper-parameters among the values detailed in the GitHub repository. The methods described in this section were implemented using the sklearn Python library.

---
[1] https://kaldi-asr.org/models/m1
[2] https://huggingface.co/jonatasgrosman/wav2vec2-xls-r-1b-english
[3] https://huggingface.co/jonatasgrosman/wav2vec2-large-xlsr-53-english
[4] https://huggingface.co/openai/whisper-base
[5] https://huggingface.co/openai/whisper-large
[6] https://github.com/LuciaGomZa/INTERSPEECH2023_AlzheimersDisease.git

*3.1.2. Word Embeddings and Neural Network*

Neural network based models have transformed the Natural Language Processing (NLP) field, significantly improving the state-of-the-art results in numerous tasks, including AD recognition from speech transcriptions [29, 30, 31]. Particularly, text embeddings provide efficient vector representations of the text without the need of the time-consuming feature extraction and selection. Here, transcriptions were lowercased, tokenized and padded to a maximum length of 250 words based on the word count distribution of the transcriptions. Then, the pre-trained FastText embedding trained on Common Crawl [32] was used to convert the transcriptions of the participants' responses into word vectors. These representations then were fed into a neural network (NN), composed of a bidirectional Long Short-Term Memory (biLSTM) layer with 128 units that returns all the hidden states. A 0.2 dropout rate was used to prevent over-fitting. After one-dimensional global max pooling, the result was fed into a dense layer of 64 units and a final binary classification layer. Due to the relatively small dataset, the results may vary depending on the model initialisation. Therefore, following the procedure proposed by [33], the model was run 25 times with different random seeds. The majority voting of the predictions was used to set the label of each sample in the test set. The neural network architecture used the following hyperparameters: binary cross entropy loss, Adam optimizer, batch size of 10 and 30 epochs. It was implemented using the Keras Python library.

### 3.2. Pause Information

To include pause information, the modelling approaches described above were adapted. For the one based on feature extraction and machine learning, 4 additional features were calculated: the speech rate, as the ratio between the number of words in the transcriptions and the audio duration; the total number of pauses, the mean length of the pauses and the summed pause durations. These features were added to the LIWC feature set, and the same pipeline described in Section 3.1.1 was applied.

With regards to word embeddings and neural network, the raw text is used directly to do the classification. To include pause information, the pause encoding procedure described in [33] was applied. First, the original punctuation of both manual and ASR-based transcriptions was removed. Timestamps from the Kaldi-based forced aligner and WhisperX, respectively, were used to calculate the pauses between the words in the full speech recordings. Finally, the following encoding was applied: short pauses (<0.5s) were replace by ","; medium pauses (0.5-2s) were replaced by "."; and long pauses (>2s) were replaced by "...". Pauses shorter than 50 ms were excluded. Once the pauses were included as punctuation marks in the transcriptions, the pipeline in Section 3.1.2 was applied.

### 3.3. Punctuation

One of the main differences between state-of-the-art ASR models using Kaldi, Wav2vec2.0 and Whisper, is that the latter includes automatic punctuation. When training the machine learning models considering punctuation, a specific LIWC category with 12 features that counts punctuation marks was included in the feature set, resulting in 93 features. As for the neural network, the raw text used as model input was analysed with and without punctuation marks.

## 4. Results

In this section, we first present the evaluation of the ASR-based transcriptions obtained with five state-of-the-art ASR models. Then, we summarise the AD classification results for the different conditions and model approaches described before.

### 4.1. WER of the ASR Models

In order to determine which ASR model we use for obtaining the automatic transcriptions, we evaluated the three different ASR systems, ASpIRE (Kaldi), Wav2vec2 and Whisper, by means of the Word Error Rate (WER), using the manual transcriptions of the training set as the ground truth. The results are shown in Table 1, where the mean value for the transcriptions in the training set is presented, as well as the mean value for AD and HC groups independently.

Table 1: *WER (%) of the ASR models in the training set.*

| ASR model | WER Mean (std) | WER HC Mean | WER AD Mean |
|---|---|---|---|
| ASpIRE | 61.43 (22.33) | 61.38 | 61.47 |
| Wav2vec2.0-large-xlsr-53 | 65.00 (17.67) | 61.58 | 68.43 |
| Wav2vec2.0-xls-r-1b | 41.11 (18.11) | 38.88 | 43.33 |
| Whisper-base | 44.02 (23.81) | 44.08 | 43.96 |
| Whisper-large | **30.18 (21.04)** | **29.05** | **31.31** |

These results show that the WER is higher in the AD cases for almost all ASR models except for one (43.96% and 44.08%, AD and HC, Whisper-base), as AD patients tend to use more filler and incomplete words. The Whisper-large model outperforms the other ASR systems for all three groups: all (WER=30.18%), HC (29.05%) and AD (31.31%). Therefore, the ASR-based transcriptions obtained with Whisper-large were used in the present study for AD classification.

### 4.2. Classification Results

Table 2 shows the classification results for the different modelling approaches. It includes the evaluation metrics in the test set, as well as the number of features selected (N) by the machine learning models. The first two columns of the table indicate whether the model inputs included pause information and punctuation marks. When studying the pauses, the transcriptions include commas and dots as explained in Section 3.2, but they correspond to the coding of pauses rather than being considered punctuation marks in the table.

Better test accuracy results are displayed in Table 2 with manual transcriptions (ranging from 0.667 to 0.854) than with ASR-based transcriptions (0.625 to 0.833). In general, higher test results are obtained with the neural network (accuracy ranging from 0.729 to 0.854) compared to the traditional machine learning approach (accuracy ranging from 0.625 to 0.830). However, for ASR-based transcriptions, the best ML result (0.830, for FE+KNN) is only slightly lower than the best NN result (0.833, for WE+NN); and both results are marginally outperformed by the best result for manual transcriptions (0.854, also for WE+NN). Including pause encoding clearly improves the results in both manual and ASR-based transcriptions. The test accuracy for the WE+NN model increases from a maximum of 0.792 without pauses to 0.854 when they are included for manual transcriptions and from 0.750 to 0.833 when using the ASR-based ones. Conversely, punctuation marks do not appear to have a significant impact on the classification performance.

Table 2: *Testing evaluation metrics for the different modeling approaches, for both manual and ASR transcriptions.*

| Pauses | Punct. | Model | Manual transcriptions | | | | | ASR-based transcriptions | | | | |
|---|---|---|---|---|---|---|---|---|---|---|---|---|
| | | | N | Acc | Precision | Recall | F1 | N | Acc | Precision | Recall | F1 |
| No | No | FE + SVM | 41 | 0.708 | 0.692 | 0.750 | 0.720 | 38 | 0.750 | 0.800 | 0.667 | 0.727 |
| | | FE + KNN | 40 | 0.667 | 0.833 | 0.417 | 0.556 | 14 | 0.646 | 0.733 | 0.458 | 0.564 |
| | | WE + NN | - | 0.792 | 0.792 | 0.792 | 0.792 | - | 0.729 | 0.789 | 0.625 | 0.697 |
| | Yes | FE + SVM | 36 | 0.667 | 0.667 | 0.667 | 0.667 | 27 | 0.667 | 0.682 | 0.625 | 0.652 |
| | | FE + KNN | 13 | 0.688 | 0.765 | 0.542 | 0.634 | 10 | 0.625 | 0.667 | 0.500 | 0.572 |
| | | WE + NN | - | 0.771 | 0.783 | 0.750 | 0.766 | - | 0.750 | 0.773 | 0.708 | 0.739 |
| Yes | - | FE + SVM | 48 | **0.771** | 0.724 | 0.875 | 0.792 | 52 | 0.809 | 0.818 | 0.783 | 0.800 |
| | | FE + KNN | 15 | 0.708 | 0.812 | 0.542 | 0.650 | 12 | **0.830** | 0.895 | 0.739 | 0.810 |
| | | WE + NN | - | **0.854** | 0.870 | 0.833 | 0.851 | - | **0.833** | 0.833 | 0.833 | 0.833 |

## 5. Discussion

The results in Section 4.1 show that the lowest WERs are obtained with the Whisper-large model, i.e. a WER of 30.18% for the training set. Therefore, we used Whisper-large transcriptions in addition to the manual transcriptions provided by the ADReSS challenge 2020 to study classification accuracy.

With regard to our first research question, *RQ1*, not surprisingly, the use of manual transcriptions led to better classification results in all cases (with or without pauses and with or without punctuation marks) since the ASR-based transcriptions of doctor-patient consultations are still far from perfect in comparison to the manual ones due to speech disfluencies, bad audio quality, background noises or simultaneous speech segments. However, the accuracy difference between the best manual value (0.854) and the best automatic one (0.833) is small.

As for our *RQ2*, we reported AD classification results of deep learning over traditional machine learning approaches over both manual and ASR-based transcriptions. The first approach is based on feature engineering and machine learning algorithms with two different classifiers SVM and KNN, in which we combined Whisper output with LIWC features, pauses and speech rate. The main advantage of this method is that results can be interpreted as we selected the features. In fact, we observed that many features selected by the feature selection method were plausible. For instance, the speech rate was selected in all the models that included pause information, as AD participants present longer recordings but shorter transcriptions compared to HC. Moreover, the "six or greater letter words" (Sixltr) category was selected in ten out of the twelve classical machine learning models tested. When compared to healthy individuals, those with cognitive deficits tend to use fewer of these words, a reasonable outcome given their condition [26].

Additionally, we studied the influence of linguistic features with FastText WE using the raw text as input. In all cases, except for the one in which we do not use pauses nor punctuation information for modelling ASR-based transcriptions, the deep leaning approach (WE+NN) outperforms the machine learning methods, being the best results for manual transcriptions 0.854 (WE+NN) and 0.771 (FE+SVM), and the best results for ASR-based transcriptions 0.833 (WE+NN) and 0.830 (FE+KNN).

Finally, regarding *RQ3*, we studied the influence of timing (pauses) information and automatic punctuation provided by the Whisper-large ASR model in order to determine their implications on AD classification in comparison to manually-transcribed data provided by human transcribers. In both cases, manual and ASR-based transcriptions, and in both learning approaches (classical machine learning and deep learning), the isolated use of punctuation marks by itself does not improve the performance of the classification system; but the use of a pause encoding procedure in the deep learning models leads increases the accuracy of the AD classification. This is inline with findings of [19], who apply a pause encoding process based on voice activity detection. However, we achieve a higher accuracy using our pause encoding process on both manual transcriptions and ASR-based transcriptions, 0.854 and 0.833 respectively, compared to the 0.83 and 0.81 achieved by [19]. Our ASR-based transcription score also outperforms the 0.761 reported by [18].

## 6. Conclusion

In this work, we have presented and experimentally evaluated a comparative study and methodology for AD classification using manual and ASR-based transcriptions. Our method leverages ASR from Whisper-large model and a combination of machine (LIWC features and SVM and KNN classifiers) and deep learning approaches (FastText WE and neural network). The experiments we conducted on the ADReSS dataset have provided evidence that the use of linguistic features from ASR-based transcriptions generated by the state-of-the-art, Whisper-large model, beside a pause encoding procedure, is a plausible approach for AD classification.

Although the traditional ML models are easier to interpret, the best results come from the deep learning approach, WE+NN, and for manual transcriptions the accuracy is higher (0.854) than with ASR transcripts (0.833). Therefore, the findings demonstrate that linguistic features, extracted from Whisper-large ASR transcripts, can deliver comparable classification results to those obtained from manual transcripts. However, there is still room for ASR improvement (e.g. finetuning the ASR model with realistic acoustic AD data). These findings suggest that automatic AD classification, solely based on speech and language features, is a promising avenue to explore.

## 7. Acknowledgements

This work was supported by the Generalitat Valenciana (ACIF/2021/187 and CIBEFP/2021/69) and by the Universitat Politècnica de València (PAID-10-20).